\documentclass[12pt]{article}
\usepackage[margin=1in]{geometry}
\usepackage{lineno}
\usepackage{graphicx}%
\usepackage{multirow}%
\usepackage{amsmath,amssymb,amsfonts}%
\usepackage{amsthm}%
\usepackage{mathrsfs}%
\usepackage[title]{appendix}%
\usepackage{xcolor}%
\usepackage{textcomp}%
\usepackage{manyfoot}%
\usepackage{booktabs}%
\usepackage{algorithm}%
\usepackage{algorithmicx}%
\usepackage{algpseudocode}%
\usepackage{listings}%
\usepackage[colorlinks=true, linkcolor=blue, citecolor=blue, urlcolor=blue]{hyperref}
\usepackage{cleveref}
\usepackage{caption}
\usepackage{subcaption}
\begin{document}

\title{Specialized Foundation Models \\for Intelligent Operating Rooms}

% \author*[1,2]{\fnm{Ege} \sur{Özsoy}}\email{ege.oezsoy@tum.de}

% \author[1,2]{\fnm{Chantal} \sur{Pellegrini}}\email{chantal.pellegrini@tum.de}

% \author[1,2]{\fnm{David} \sur{Bani-Harouni}}\email{david.bani-harouni@tum.de}

% \author[1,2]{\fnm{Kun} \sur{Yuan}}\email{kun.yuan@tum.de}

% \author[1,2]{\fnm{Matthias} \sur{Keicher}}\email{matthias.keicher@tum.de}

% \author[1,2]{\fnm{Nassir} \sur{Navab}}\email{nassir.navab@tum.de}

% \affil*[1]{\orgdiv{Computer Aided Medical Procedures}, \orgname{Technical University Munich}}
% \affil[2]{\orgdiv{Munich Center for Machine Learning (MCML)}, \orgname{Germany}}

\author{
  Ege Özsoy\textsuperscript{1,2},
  Chantal Pellegrini\textsuperscript{1,2},
  David Bani-Harouni\textsuperscript{1,2},\\
  Kun Yuan\textsuperscript{1,2},
  Matthias Keicher\textsuperscript{1,2},
  Nassir Navab\textsuperscript{1,2}
}

\date{}

\maketitle

\begin{center}
\texttt{ege.oezsoy@tum.de} \\
\textsuperscript{1}Munich Center for Machine Learning (MCML), Germany \\
\textsuperscript{2}Computer Aided Medical Procedures, Technical University of Munich \\
\vspace{1em}
\end{center}

\abstract{\textit{Surgical procedures unfold in complex environments demanding coordination between surgical teams, tools, imaging and increasingly, intelligent robotic systems. Ensuring safety and efficiency in ORs of the future requires intelligent systems, like surgical robots, smart instruments and digital copilots, capable of understanding complex activities and hazards of surgeries. Yet, existing computational approaches, lack the breadth, and generalization needed for comprehensive OR understanding. We introduce ORQA, a multimodal foundation model unifying visual, auditory, and structured data for holistic surgical understanding. ORQA’s question-answering framework empowers diverse tasks, serving as an intelligence core for a broad spectrum of surgical technologies. We benchmark ORQA against generalist vision-language models, including ChatGPT and Gemini, and show that while they struggle to perceive surgical scenes, ORQA delivers substantially stronger, consistent performance. Recognizing the extensive range of deployment settings across clinical practice, we design, and release a family of smaller ORQA models tailored to different computational requirements. This work establishes a foundation for the next wave of intelligent surgical solutions, enabling surgical teams and medical technology providers to create smarter and safer operating rooms.}}

\maketitle

\begin{figure}[ht]
\centering
\includegraphics[width=0.85\linewidth]{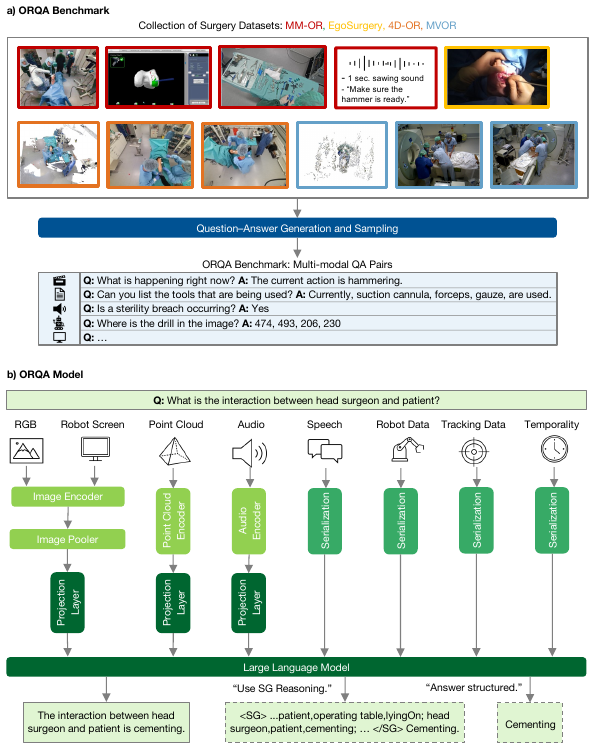}
\caption{Overview of the ORQA benchmark and model. Four OR datasets are unified into a multimodal QA benchmark. Our model encodes samples via modality-specific encoders and processes them with an LLM, optionally leveraging scene graphs~(SG) for reasoning, and outputs structured or free-text responses.} \label{fig:model}
\end{figure}

Operating rooms (ORs) are among the most complex and high-risk environments in medicine, where intelligent systems are increasingly essential to ensure safety, efficiency, and coordination~\cite{protserov2024development, kanji2021work}. Across a wide range of procedures, from routine laparoscopic cholecystectomies to robot-assisted joint replacements, surgical teams and robotic systems must coordinate dynamically in confined spaces, making split-second decisions based on continuously evolving multimodal information. This information includes visual cues, auditory signals, patient vitals, instrument telemetry, and even haptic feedback in robot-assisted interventions. Maintaining a shared, real-time understanding of the surgical scene, what instruments are in use, where attention is focused, and which anatomical structures are at risk, is critical for avoiding complications and guiding the next steps in the procedure.

The increasing integration of digital tools, robotic assistance, and sensor-rich platforms into the OR has dramatically increased both the quantity and complexity of intraoperative data, heightening the need for computational systems capable of real-time interpretation and reasoning, to support real-time decision-making, anticipate procedural steps, and facilitate coordinated surgical workflows~\cite{lalys2014,maier2017surgical}.

Over the past decade, advances in surgical data science have yielded a variety of specialized algorithms aimed at discrete subproblems: surgical phase recognition~\cite{twinanda2016endonet,jin2020multi}, which segments a procedure into predefined steps; action detection~\cite{rendezvous}, which identifies fine-grained gestures such as “cut”, “suturing”, or “retraction”; instrument detection and segmentation~\cite{ding2022carts,pei2024s,guo2024tri}, which localize and classify tools; and surgical visual question answering~\cite{he2024pitvqa,yuan2024advancing}, enabling models to answer single, pre-specified questions about intraoperative scenes. However, most efforts have focused on narrow objectives using limited datasets and modalities, and are often restricted to internal surgical views. 

More recent work has broadened the scope of surgical AI to include external perspectives and richer representations. Multi-view RGB-D datasets such as MVOR~\cite{mvor} capture human poses from external, ceiling mounted cameras, enabling pose detection in the OR. EgoSurgery~\cite{fujii2024egosurgery} provides egocentric video and gaze data from surgeons, facilitating the study of visual attention and hand-tool coordination. TeleOR~\cite{wu2024teleor} reconstructs real-time 3D models of the OR scene to support remote collaboration and automated documentation. Scene graph–based approaches, such as 4D-OR~\cite{ozsoy20224d}, encode semantic relationships among people, tools, and patient anatomy, based on multi-view camera inputs, offering structured abstractions of OR activities~\cite{ozsoy2023,pei2024s,guo2024tri,ozsoy2024oracle, ozsoy2025mmor}. While these efforts represent significant steps toward holistic OR modeling, each remains focused on one or two tasks like pose estimation, semantic segmentation, or structured scene representation, with limited cross-task generalization and dataset scope.

In parallel, the broader field of biomedical AI has witnessed the rise of foundation models: large, pretrained vision-language or multimodal architectures that demonstrate impressive zero-shot or few-shot performance across a range of clinical tasks. In radiology, models leverage large-scale image-text datasets to generate reports, answer diagnostic questions, and assist in anomaly detection across chest X-rays and CT scans~\cite{wu2023towards,pellegrini2023radialog}. In pathology, multimodal LLMs are trained on histopathology slides and associated diagnostic notes to provide interpretive guidance and triage~\cite{lu2024multimodal}. In endoscopy, large vision-language frameworks such as LLaVA-Surg fuse video streams and clinical text to detect polyps and guide therapeutic decisions in real time~\cite{li2024llavasurg}. Despite these successes, no comparable foundation model exists for the operating room: a domain with uniquely high stakes, dense sensory inputs, and rapidly shifting contexts that combine video, audio, device telemetry, and human interactions.

This is not primarily caused by a lack of conceptual understanding. Indeed, large language models have been exposed to surgical terminology, procedure descriptions, and basic medical protocols through their extensive text pretraining. When prompted appropriately, they can often articulate what constitutes a sterility breach, or describe the steps of a laparoscopic cholecystectomy. However, their weakness lies in visual understanding. The visual inputs encountered in operating rooms, multi-view RGB-D feeds, egocentric and exocentric perspectives and complex spatial layouts, and synchronized robot kinematics or audio, are fundamentally different from the curated Internet-scale datasets these models are trained on. Consequently, even the most capable generalist models struggle to extract clinically meaningful signals from OR data. They may recognize that “a surgeon is present,” yet fail to localize a specific instrument, detect a handoff event, or infer that a robot is improperly docked. The difficulty is further amplified by the lack of alignment between their implicit visual representations and the structured supervision signals required for precise, task-specific reasoning in the OR, such as understanding surgical phase transitions, scene graph relations, or spatial constraints. In short, while these models demonstrate strong capabilities in language-based medical tasks, their visual understanding remains poorly suited to the complex, multimodal demands of surgical environments.

To quantify these limitations, we evaluate two leading proprietary generalist vision-language foundation models, ChatGPT and Gemini, as well as an open source vision-language model, Qwen2-VL, on our newly developed multimodal OR evaluation suite. Despite providing detailed prompts, multiple visual inputs, and surgery-specific context, each model’s performance on our benchmark hover only slightly above a trivial statistical baseline, revealing severe deficiencies in spatial reasoning, temporal forecasting, and safety-related inferences. These findings underscore the necessity of training models on domain-relevant, structured surgical data.

In light of these limitations, we make the following contributions: \textbf{First}, we introduce the ORQA (Operating Room Question Answering) benchmark, a comprehensive multimodal benchmark designed to evaluate surgical scene understanding through a unified question-answering (QA) framework. ORQA unifies annotations from four publicly available OR datasets, MVOR~\cite{mvor}, 4D-OR~\cite{ozsoy20224d}, EgoSurgery~\cite{fujii2024egosurgery}, and MM-OR~\cite{ozsoy2025mmor}, and defines 23 diverse tasks covering spatial reasoning, temporal dynamics, procedural understanding, and safety-critical tasks. By framing these tasks uniformly as QA pairs, the ORQA benchmark enables model comparison across modalities, datasets, and task types using a single, interpretable output format. \textbf{Second}, we evaluate proprietary state-of-the-art vision-language foundation models and open source models on our benchmark in a zero-shot setting and show their limited performance, highlighting the shortcomings of current vision-language systems in handling the specialized visual and multimodal signals encountered in surgery. \textbf{Third}, we develop the ORQA model, an OR-specialized foundation model, trained on one million curated QA pairs sampled from our benchmark. It integrates diverse input modalities, including video, point clouds, metadata, and audio, through modality-specific encoders and a large language model backbone. The resulting model significantly outperforms generalist models by a wide margin and demonstrating strong performance across spatial, temporal, and safety-critical tasks, showing that foundation model approaches are viable in surgical domains, but only when trained with surgical data. Further, to support deployment in resource-constrained environments, we used knowledge distillation to create smaller and faster variants of ORQA. These models maintain high task accuracy while offering significant reductions in inference time and memory usage, enabling local deployment on real-time surgical hardware.

Together, our contributions establish a foundation for building and evaluating the next generation of intelligent surgical systems, including autonomous robotics, real-time decision support, and automated documentation.
\begin{table}[t]
\centering
\footnotesize
\caption{ORQA tasks and their expected answers.}
\label{tab:orqa_tasks}
\begin{tabular}{l|l}
\toprule
\textbf{Task}       & \textbf{Expected Answer} \\ \midrule
People Counting     & Number of people in the OR \\ 
Role Detection      & List of roles of people in the OR \\
Interaction Detection      & Interaction between $Entity_i$ and $Entity_j$ \\
Attribute Detection      & $Attribute_x$ of $Object_i$ e.g. color of drill \\
Action Detection      & Name of the current action \\
Estimate Time Until  & Number of seconds until $Action_x$ \\
Estimate Status      & Progress in percent of current action \\
Is Completed  & Binary indication if $Action_x$  was already performed\\
Is Base Array Visible     & Binary indication if robot base array is visible \\
Is Robot Calibrated      & Binary indication if robot is calibrated \\
Sterility Breach Detection      & Binary indication, if sterility breach is happening \\
Robot Step Detection      & Name of the current robot step \\
Next Robot Step Estimation      & Name of the next robot step \\
2D Detection      & Bounding box coordinates of $Entity_i$ in the image \\
3D Detection      & 3D Center point of $Entity_i$ in the OR \\
3D Distance      & Distance between $Entity_i$ and $Entity_j$ in meters \\
Tool Detection      & List of tools currently used \\
Scene Graph Generation      & Current scene graph as list of triplets \\
Entity Detection      & List of all current entities in the OR \\
Sorted Entity Detection & List of all current entities in the OR from left to right \\
Gaze Location      & Coordinates of the gaze of the surgeon \\
Gaze Object Detection      & Name of the object the surgeon is looking at \\
Monitor Text OCR      & Summary of the information on the monitor \\

\bottomrule
\end{tabular}
\end{table}

\begin{figure}[tb]
\centering
\includegraphics[width=0.6\textwidth]{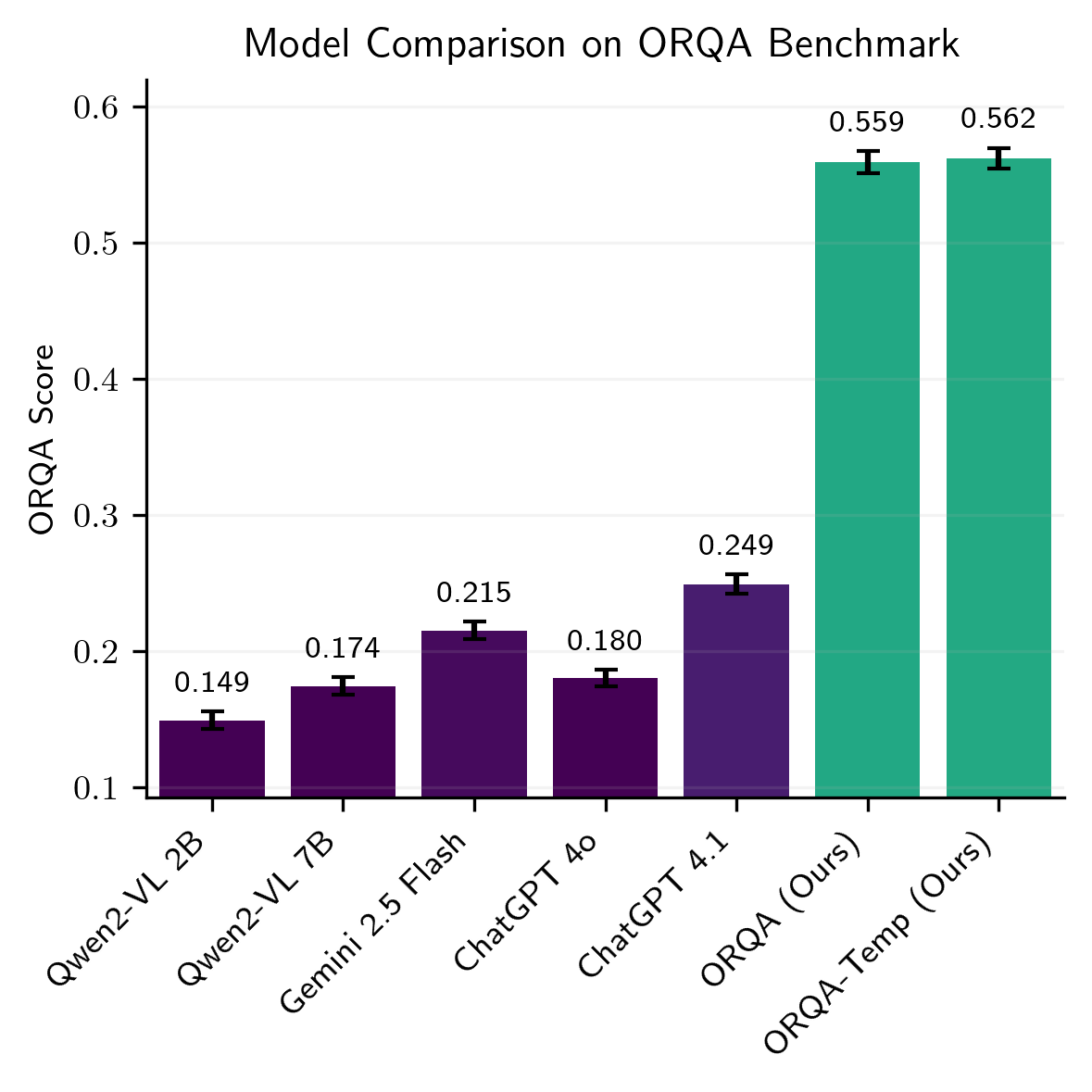}
\caption{Comparison of the domain-specialized ORQA model with state-of-the-art generalist vision-language foundation models on our benchmark test set. We report the ORQA Score along with 95\% confidence intervals.}
\label{fig:main_results}
\end{figure}

\begin{figure}[tb]
\centering
\begin{minipage}{0.65\textwidth}
    \centering
    \includegraphics[width=\textwidth]{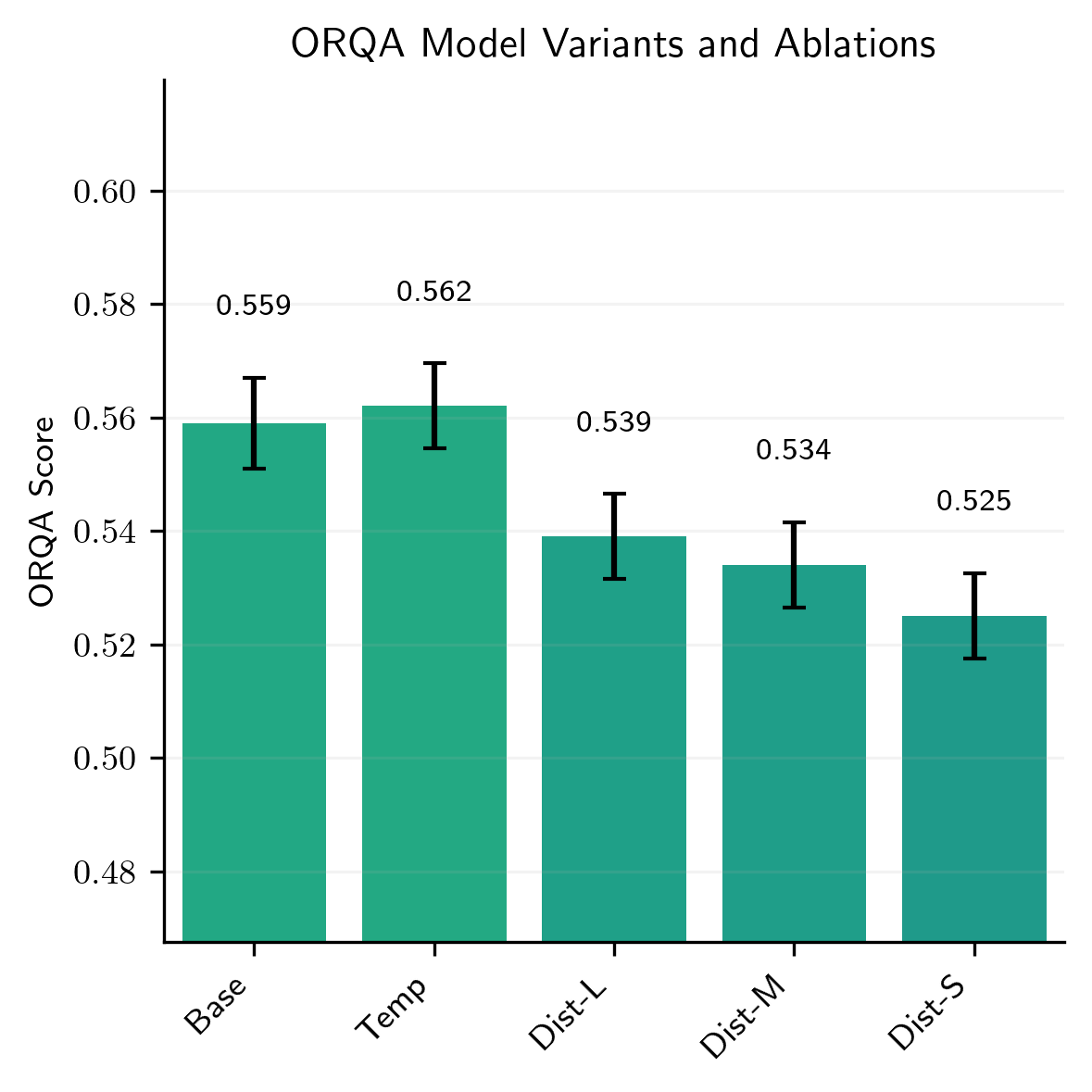}
\end{minipage}

\vspace{0.8em}  % Adjust spacing between figure and table

\begin{minipage}{\textwidth}
    \centering
    \footnotesize
    \begin{tabular}{l|c|c|c|c|c}
        \toprule
        \textbf{Model} & \#\textbf{Layers} & \textbf{Hidden Size} & \textbf{\#Params} & \textbf{T/s} & \textbf{ORQA Score (95\% CI)} \\
        \midrule
        Base     & 28 & 1536 & 1.78B & 98  & 0.559 [0.551, 0.567] \\
        Temp     & 28 & 1536 & 1.78B & 88  & 0.562 [0.553, 0.568] \\
        \midrule
        Dist-L   & 28 & 768  & 512M  & 144 & 0.539 [0.530, 0.545] \\
        Dist-M   & 15 & 768  & 360M  & 218 & 0.534 [0.526, 0.541] \\
        Dist-S   & 8  & 768  & 278M  & 319 & 0.525 [0.517, 0.532] \\
        \bottomrule
    \end{tabular}
\end{minipage}

\vspace{0.5em}
\caption{Performance and efficiency trade-offs across ORQA model variants and ablations. The bar chart shows model performance with 95\% confidence intervals. The table below details architectural specifications and inference throughput (tokens per second). Distilled variants offer smaller sizes and higher throughput with moderate performance degradation.}
\label{fig:orqa_variants_combined}
\end{figure}

\begin{figure}[tb]
\centering
\includegraphics[width=0.85\linewidth]{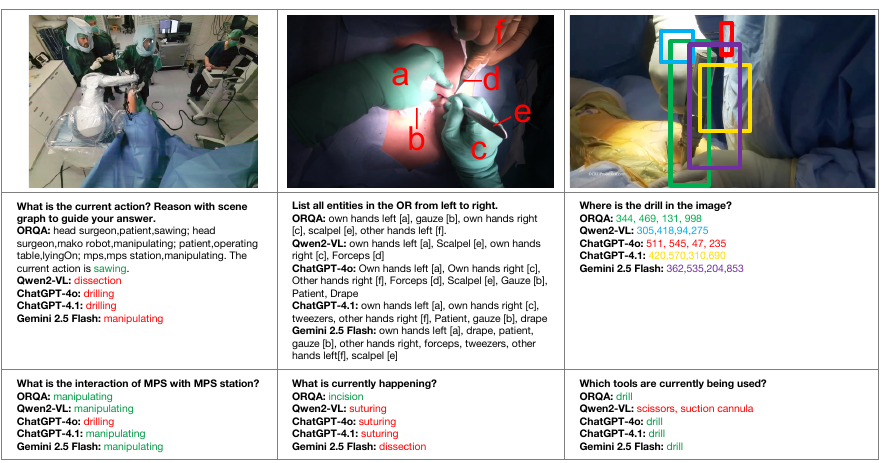}
\caption{Qualitative examples from (left to right) MM-OR, EgoSurgery and a zero-shot image from the internet. For all non ORQA models, we explicitly prompt with possible answers per question and the expected output formats, and manually parse their answers. The ground truth order of objects (a-f) and the predicted bounding boxes of all models are visualized in image two and three respectively. For all other questions, correct answers are marked in green, incorrect ones in red.}
\label{fig:qualitative}
\end{figure}

\begin{figure}[tb]
\centering
\includegraphics[width=0.85\linewidth]{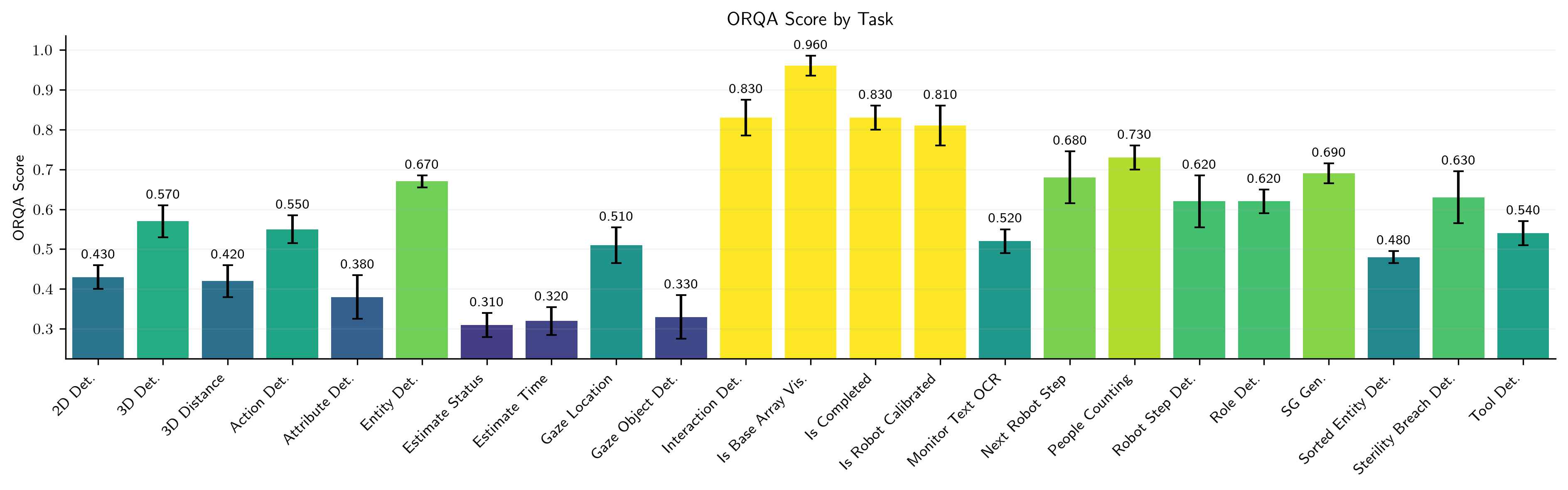}
\caption{Performance of ORQA across all 23 clinical tasks in the ORQA benchmark. The ORQA Score, reported per task, captures accuracy across diverse challenges, including spatial reasoning, workflow prediction, and safety monitoring. Results demonstrate robust performance and broad generalizability of domain-specialized pretraining for surgical scene understanding.}

\label{fig:orqa_score_by_task_transposed}
\end{figure}

\begin{figure}[tb]
    \centering
    \includegraphics[width=0.85\linewidth]{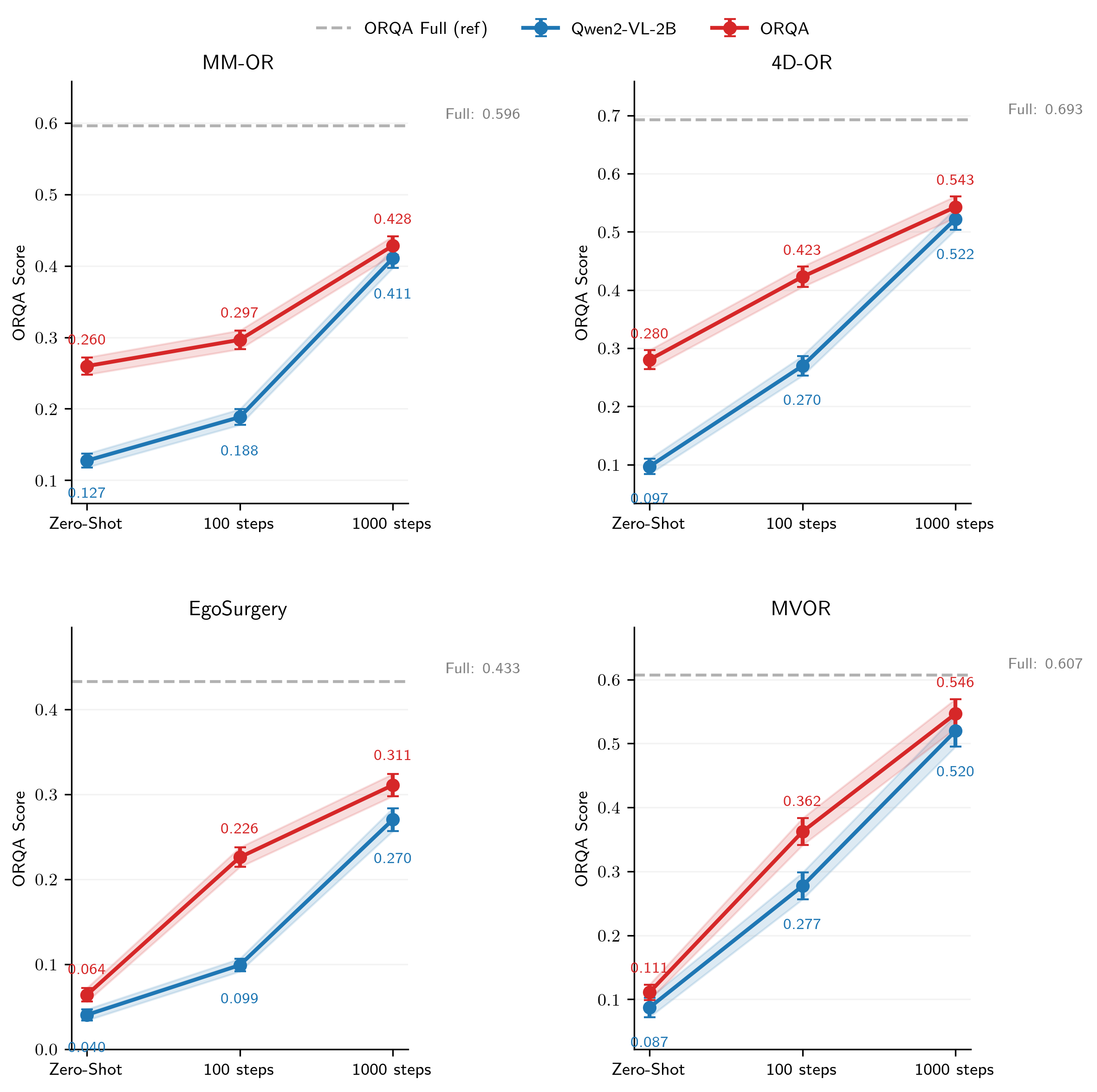}
        % \begin{subfigure}[b]{0.4\textwidth}
    %     \includegraphics[width=\textwidth]{figures/figure4_leaveoneout_mmor.png}
    % \end{subfigure}
    % \begin{subfigure}[b]{0.4\textwidth}
    %     \includegraphics[width=\textwidth]{figures/figure5_leaveoneout_4dor.png}
    % \end{subfigure}

    % \begin{subfigure}[b]{0.4\textwidth}
    %     \includegraphics[width=\textwidth]{figures/figure6_leaveoneout_egosurgery.png}
    % \end{subfigure}
    % \begin{subfigure}[b]{0.4\textwidth}
    %     \includegraphics[width=\textwidth]{figures/figure7_leaveoneout_mvor.png}
    % \end{subfigure}
    \caption{Leave-one-out generalization experiments. We evaluate how well ORQA and its base model, Qwen2-VL-2B, generalize to unseen datasets by systematically leaving out one dataset during training and evaluating on it. For each held-out dataset, we report four configurations: zero-shot performance without any exposure to the dataset, performance after fine-tuning on it for 100 and 1000 steps, and the original ORQA model trained on all datasets, including the held-out one, shown as a reference.}
    \label{fig:leaveout_generalization}
\end{figure}

\section*{Results}

\subsection*{A Benchmark to Evaluate Holistic OR Intelligence and Perception}
To accurately evaluate the comprehension of AI systems in the OR, we create the ORQA benchmark, a comprehensive multimodal question-answering dataset that unifies annotations from four public OR datasets, MVOR~\cite{mvor}, 4D-OR~\cite{ozsoy20224d}, EgoSurgery~\cite{fujii2024egosurgery}, and MM-OR~\cite{ozsoy2025mmor}. ORQA supports 23 clinically relevant tasks, including spatial reasoning (e.g., ``What is the distance between the surgeon and the tool cart?''), procedural anticipation (e.g., ``What is the next robot step?''), and safety monitoring (e.g., ``Is there a sterility breach?''). The full taxonomy of tasks and expected answers is shown in~\Cref{tab:orqa_tasks}.

Given the raw annotations from each dataset, we generate over 100 million QA pairs, then curate this into a training set of 1 million high-diversity samples using a task-aware sampling strategy that maximizes answer diversity and contextual coverage (details in~\nameref{sec:method}). For validation and testing, we analogously create 10,000 QA pairs each. This unique benchmark serves a dual role: (1) as a standardized evaluation suite for multimodal OR models, and (2) as a training source that enables large models to develop a broad and structured understanding of surgical environments.

By framing OR scene understanding as a large-scale question-answering problem, the ORQA benchmark offers several advantages, such as being able to rely on one unified model architecture for solving many critical tasks, and facilitating a standardized way to compare models across a wide array of vision, reasoning, and safety tasks that mirror real-world intraoperative challenges.

\subsection*{ChatGPT and Similar Generalist Models Struggle to Grasp the Operating Room}
To assess whether existing foundation models can generalize to the OR setting, we evaluate three proprietary state-of-the-art vision-language models, ChatGPT-4o and 4.1, Gemini 2.5 Flash, and two open source vision-language model Qwen2-VL 2B and 7B, in a zero-shot fashion on the ORQA benchmark. These models are tested using detailed task-specific prompts that include question formatting, description of the question, expected outputs, and relevant OR context. Performance is reported using the ORQA Score metric (0–1), which accounts for task-specific evaluation criteria (see~\nameref{sec:method}). As shown in~\Cref{fig:main_results}, despite this careful prompt engineering, the generalist models performed only marginally better than a trivial statistical baseline, which always predicts the \textbf{most frequent answer} for each task (e.g., ``no'' for binary predicates, the average 3D distance for distance queries) and achieves a score of 0.191 (95\% CI: [0.185, 0.199]). GPT-4o, 4.1 and Gemini achieve overall scores of 0.180 (95\% CI: 0.174–0.186), 0.249 (95\% CI: 0.242–0.256) and 0.215 (95\% CI: 0.209–0.222), respectively, while Qwen2-VL 2B and 7B reach 0.149 (95\% CI: 0.142–0.155) and 0.174 (95\% CI: 0.167–0.180). A detailed breakdown by tasks reveals these models struggle with all tasks, but particularly so on spatial and temporal reasoning tasks and frequently hallucinate answers when confronted with domain-specific concepts such as robot steps or sterility protocols. Qualitative results, shown in~\Cref{fig:qualitative}, further highlight these limitations. For instance, when asked “\textit{Where is the drill in the image?}” on an internet sample, all models except Gemini 2.5 Flash and ORQA predict bounding boxes that are spatially distant from the ground truth, sometimes overlapping with irrelevant instruments or background structures. On the MM-OR and EgoSurgery frames, only ORQA correctly identifies the current action.

These findings underscore that the surgical operating room presents unique multimodal challenges, dense visual clutter, specialized instruments, context-dependent actions, and precise safety constraints, that current generalist foundation models are ill-equipped to handle. Despite their strong performance in other biomedical domains, their training on broad web-scale or general medical corpora and lack of exposure to surgical-specific data, tasks, and workflows fails to cover the structured, procedural, and safety-critical knowledge required for OR comprehension.

\subsection*{Specialized Surgical Models Significantly Outperform ChatGPT and Other General Foundation Models}

Having established the limitations of generalist vision-language models, we develop and evaluate ORQA, a domain-specialized multimodal foundation model explicitly trained on surgical data spanning diverse modalities, hospitals, and tasks. Unlike ChatGPT, Gemini, or Qwen2-VL, ORQA is trained on QA pairs derived from structured surgical annotations, enabling it to learn clinically meaningful patterns of interaction, spatial reasoning, and procedural logic. Further it can explicitly integrate modalities such as point clouds, audios and more. As shown in~\Cref{fig:main_results}, ORQA model achieves a significantly higher average score of 0.559 (95\% CI: [0.551, 0.567]), outperforming all baselines by a large margin. Notably, as shown in~\Cref{fig:orqa_score_by_task_transposed}, ORQA exhibits good performance across a diverse set of tasks, modalities, and datasets, underscoring the strength of domain-specific pretraining. In contrast to the erratic outputs of generalist models (\Cref{fig:qualitative}), ORQA produces consistent, accurate and structured responses. For example, it can reason about spatial proximity in 3D point clouds, infer the sequence of robotic steps from prior interactions, assess sterility breaches based on auditory and visual cues and localize tools accurately, even on out of distribution surgical data publicly available on the internet. We further extend ORQA to ORQA-Temp, where the model uses memory scene graphs~\cite{ozsoy2023}, with the aim of leveraging surgical workflow continuity for tasks that require temporal reasoning. This approach yields only a modest performance boost, suggesting that further gains from temporal modeling may require fundamentally different or more advanced techniques.

To further assess ORQA's utility as a foundation for surgery understanding, we conducted leave-one-out experiments (\Cref{fig:leaveout_generalization}) by holding out each of the four constituent datasets during training and evaluating both ORQA and its Qwen2-VL-2B base model on the excluded dataset. For each case, we report: (1) zero-shot performance, (2) few-shot adaptation after 100 and 1000 fine-tuning steps, and (3) the original ORQA model trained on all data. Across all datasets (MM-OR, 4D-OR, EgoSurgery, MVOR), ORQA outperforms the base model in both zero- and few-shot settings, highlighting the value of domain-specific pretraining. However, zero-shot performance remains moderate, underscoring the challenge of generalizing to unseen annotation conventions. Notably, unlike Qwen2-VL, no dataset-specific prompt or structured context (e.g., predicate lists, tool definitions) is provided to ORQA. These results suggest that even minimal adaptation could significantly improve cross-dataset performance, and they establish ORQA as a strong foundation for transfer to new surgical settings.

Overall, our results demonstrate that targeted training on diverse OR-specific data enables large multimodal models to develop structured, clinically relevant perception capabilities. While full generalization across surgeries, ORs, institutions, and procedures remains an open challenge, ORQA offers a strong foundation for such transfer, especially when paired with lightweight adaptation strategies.

\subsection*{Distilling Surgical Intelligence Enables Efficient Real-Time OR Modeling}
While the full ORQA model delivers state-of-the-art performance on surgical scene understanding, its computational footprint with over 1.7 billion parameters, may pose a barrier to real-world OR integration, where hardware often consists of smaller GPUs, embedded systems, or compute-limited edge devices. To address deployment in resource-constrained environments, we employ knowledge distillation to compress ORQA into three smaller and faster variants, ORQA-Dist-L, ORQA-Dist-M, and ORQA-Dist-S, optimized for different speed and memory requirements while maintaining high task accuracy. As shown in~\Cref{fig:orqa_variants_combined}, ORQA-Dist-S achieves a $3.3\times$ speedup over the full model while maintaining an ORQA Score of 0.525 (95\% CI: [0.517, 0.532]). ORQA-Dist-M and ORQA-Dist-L offer intermediate tradeoffs with $2.2\times$ and $1.5\times$ speedups, and scores of 0.534 (95\% CI: [0.526, 0.541]) and 0.539 (95\% CI: [0.530, 0.545]), respectively. These variants reduce the parameter count by up to $6.4\times$, making them suitable for edge computing or integration into intraoperative systems with strict latency constraints.

Taken together, we show a method to effectively scale down surgical foundation models for real-time OR applications, without significant loss in performance, maintaining clinically relevant accuracy while operating within stringent computational budgets. In stark contrast to most closed source generalist systems, which often require cloud-based execution and carry potential privacy or latency risks, ORQA can be integrated locally, reducing dependence on external computation. Additionally, fast, memory-efficient models open the door to simultaneous inference across multiple OR stations or retrospective processing of entire procedures.
\section*{Discussion}
Our findings highlight a clear trajectory toward \textbf{clinically useful} surgical AI intelligence. Key considerations for real-world adoption include interpretability, efficiency, and extensibility. Surgical decision-making is inherently high-stakes, where misinterpretations can lead to complications or patient harm. ORQA’s support for \textit{structured, parsable outputs} and optional \textit{scene-graph grounding} ensures that each prediction can be traced back to underlying visual and semantic elements. For instance, when ORQA detects a \textit{sterility breach}, it can highlight the exact instrument and location, providing immediate feedback to circulating nurses. This level of transparency is crucial for surgical checklists, automated documentation, and auditing postoperative outcomes. Modern ORs may not always have the luxury of a full-scale GPU server. Portable AI assistants, embedded in ceiling-mounted cameras, laparoscopic towers, or mobile carts, require efficient inference within \textless200 ms per frame to be clinically useful. ORQA variants like ORQA-Dist-S demonstrate that it is feasible to run a 278M-parameter model at over \textit{300 tokens/sec} on a single A40 GPU, and likely \textgreater30 fps on lower-power edge devices with TensorRT or ONNX optimizations. These performance characteristics open the door to real-time surgical assistance. Furthermore, ORQA’s benchmark and architecture serve as a blueprint for future surgical foundation models. New modalities, such as haptic force sensors, intraoperative ultrasound, or wearable biosensors, can be easily integrated by extending the QA templates, augmenting the training set, and extending the multimodal encoder. As surgical robotics evolve (e.g., multi-arm platforms, autonomous suturing), ORQA’s framework can adapt by incorporating additional tasks(e.g., ``Is the suture tension within safe limits?'' or ``Which robotic arm is primary?''), enabling continuous expansion of its clinical coverage. As surgical environments and data conventions evolve, practical deployment may benefit from lightweight adaptation strategies, such as knowledge guidance~\cite{ozsoy2024oracle} or modular adapters, to further ease ORQA’s extension to novel settings with minimal supervision. Looking beyond intraoperative assistance, ORQA sets the stage for end-to-end \textit{surgical workflow intelligence}, proposing a framework for bridging preoperative planning (e.g., identifying patient-specific anatomy), intraoperative execution (e.g., monitoring instrument positions, guiding actions), and postoperative assessment (e.g., summarizing actions, predicting complications). By providing a unified, multimodal foundation, ORQA can serve as the backbone for advanced surgical care ecosystems that integrate with electronic health records (EHRs), preoperative imaging, or augmented reality.

Deploying foundation models in surgery also requires careful consideration of ethical and regulatory factors. ORQA’s reliance on publicly available, de-identified data is a step toward transparent and reproducible development, however future clinical deployments should also consider patient privacy, data governance, and potential biases, such as differences in instrument sets, workflows, and visual environments across institutions. Crucially, all ORQA variants are capable of running fully locally, enabling inference to be performed entirely within hospital infrastructure without transferring patient data to external servers. This makes ORQA compatible with privacy-preserving deployment scenarios and national data sovereignty regulations. From an operational perspective, real-world adoption will also depend on usable interfaces and robust fail-safes; here, the interpretability afforded by ORQA’s structured outputs and scene-graph-based explanations becomes especially important for fostering clinical trust and ensuring that model decisions can be audited and validated.

In summary, ORQA’s performance, efficiency, and extensibility position it as a promising foundation for surgical intelligence systems. We demonstrated that domain-specialized multimodal models can outperform generalist AI, while at the same time improving interpretability, real-time capabilities, and local deployability. This opens a path toward next-generation OR platforms that are powerful, robust and efficient.
\section*{Method}
\label{sec:method}
\subsection*{Data Sources}

The ORQA benchmark is constructed from four multimodal datasets that capture diverse surgical procedures, imaging modalities, and data collection conditions: MVOR~\cite{mvor}, 4D-OR~\cite{ozsoy20224d}, EgoSurgery~\cite{fujii2024egosurgery}, and MM-OR~\cite{ozsoy2025mmor}. Each dataset contributes distinct modalities and annotations that are harmonized into a unified representation used for question–answer (QA) generation and model training.

\noindent\textbf{MVOR} provides 732 annotated timepoints captured from ceiling-mounted RGB-D cameras in a real operating room. Each timepoint includes multi-view RGB images, depth maps, and 2D/3D keypoints for each staff member. To support temporally grounded tasks, we segmented the continuous sequences into 26 discrete clips, yielding clip-level temporal context, and manually annotated each timepoint with a surgical phase labels.

\noindent\textbf{4D-OR} includes 6,734 timepoints from simulated knee replacement surgeries, each with six RGB images, a point cloud, semantic scene graphs, clinical role annotations (e.g., ``surgeon'', ``nurse''), and panoptic segmentations for three views. We enhanced it with synthetic data inspired by ORacle~\cite{ozsoy2024oracle}, introducing variations in tool colors and equipment configurations. 

\noindent\textbf{EgoSurgery} contains 15,437 timepoints recorded from head-mounted cameras worn by surgeons in real procedures. The dataset includes egocentric video, tool bounding boxes, gaze fixation points, and manually annotated surgical phase labels.

\noindent\textbf{MM-OR} includes 92,983 timepoints from simulated robotic knee replacements performed by expert surgeons on realistic sawbones, offers high-fidelity multimodal data: multi-view RGB-D images, high-resolution detail views, audio recordings, speech transcripts, real-time robotic logs (e.g., phase, action), infrared tracking data (tool translations/rotations), 25,277 scene graphs, and panoptic segmentations for three views. We augmented MM-OR with ORacle-inspired \cite{ozsoy2024oracle} synthetic data for attribute diversity.

\subsection*{Question–Answer Generation and Sampling}
The ORQA benchmark is created by generating and curating question-answer (QA) pairs from the raw annotations of MVOR~\cite{mvor}, 4D-OR~\cite{ozsoy20224d}, EgoSurgery~\cite{fujii2024egosurgery}, and MM-OR~\cite{ozsoy2025mmor}, spanning 23 clinically relevant tasks (Table~\ref{tab:orqa_tasks}). We generate 100 million QA pairs from the training splits of the four datasets, leveraging their modality-specific annotations to cover all tasks. Automated templates are designed for each task, aligned with available data. For example, ``3D Distance'' questions (e.g., ``What is the distance between the surgeon and the tool cart?'') are derived from point clouds by computing Euclidean distances between entity centroids identified in 3D tracking data. ``Sterility Breach Detection'' questions (e.g., ``Is there a sterility breach?'') use MM-OR’s downstream task annotations to flag interactions involving non-sterile entities (e.g., (surgeon, non-sterile tool, touching)). ``Gaze Object Detection'' questions (e.g., ``What is the surgeon looking at?'') map EgoSurgery’s gaze coordinates to tool bounding boxes. 4D-OR and MM-OR, which provide scene graphs, support relational queries like “What is the interaction between the head surgeon and the patient?” or “Is the nurse currently assisting the surgeon?”. Temporal annotations, e.g., phase labels and robot state transitions, are used to construct procedure-aware questions like “What is the next robot step?” or “How long has this phase lasted?”. 

To create a robust and challenging benchmark, and avoid filling it with only trivial answers, we applied diversity sampling to reduce the 100 million QA pairs to 1 million. For each dataset, we grouped QA pairs by task and subsampled them using inverse question and answer frequency, yielding a diversity-aware dataset, where QA pairs with uncommon answers are prioritized during sampling, emphasizing long-tail patterns. 

\subsection*{Model Architecture}
The ORQA model builds on the MM2SG architecture introduced in MM-OR~\cite{ozsoy2025mmor}, extending it into a full multimodal foundation model for general-purpose operating room (OR) understanding. ORQA consists of a set of modality-specific encoders, to convert heterogeneous inputs like images, point clouds, audio, speech, robotic logs, tracker data, and temporal context, into a unified sequence of embeddings, which are processed jointly by a large language model, to generate answers to surgical queries.

\noindent\textbf{Multi-Modality Encoders.} Each modality is handled by a dedicated encoder. Images from multiple RGB-D camera views, robotic console recordings, and external tracking devices are passed through the CLIP vision encoder of Qwen2-VL, and subsequently pooled into a fixed-length sequence of $N$ tokens using a transformer-based image summarizer~\cite{ozsoy2024oracle} trained end-to-end. Point clouds are encoded via a Point Transformer V3~\cite{wu2024point}, producing a compact representation of 3D surgical environments, again trained end-to-end. Audio clips are transformed into embeddings using a frozen CLAP encoder~\cite{elizalde2023clap}, while speech transcripts are obtained from a pretrained automatic speech recognition system~\cite{radford2023robust} and directly included as text, providing the last five spoken sentences. Real-time robot logs are automatically parsed for high-level state indicators (e.g., phase, action, tool usage) and represented as structured textual summaries. Infrared tracking data, providing tool translations and rotations, are also formatted as structured text. Following previous works~\cite{ozsoy2023,ozsoy20224d}, in ORQA-temp, temporal information is provided in two tiers: (1) short-term memory using a rolling buffer of recent scene graph predictions, and (2) long-term summaries via unique historical triplets, allowing the model to retain procedural context over extended intervals.

\noindent\textbf{LLM.} All modality-specific representations are projected into the language embedding space using learned linear layers, concatenated, and provided as input tokens to the large language model. We choose to build upon Qwen2-VL~\cite{wang2024qwen2}, for its strong performance and improved multimodal compatibility. The LLM autoregressively generates an answer, considering all the modalities and the question. Depending on the prompt, our model can generate free-text strings (e.g., “The scrub nurse is holding the grasper”) or structured outputs (e.g., a list of triplets or bounding box coordinates), and can produce scene graphs before answering, enhancing explainability by grounding the answer on a summary of all relevant entities and interactions in the OR.

\begin{figure}[tb]
\centering
\includegraphics[width=0.5\linewidth]{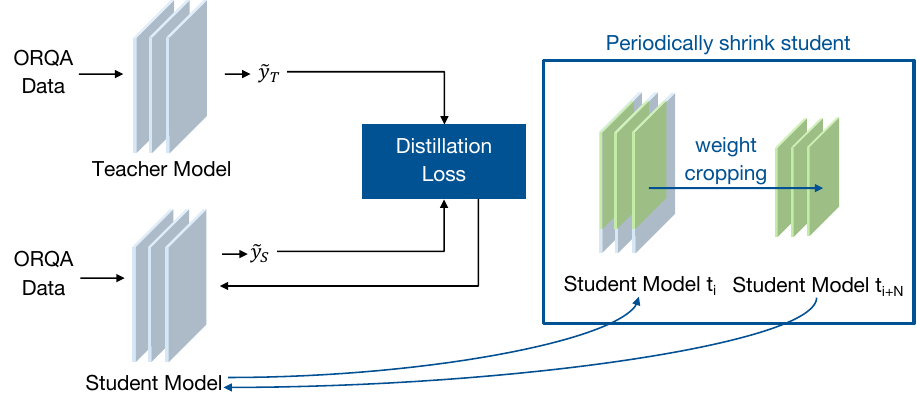}
\caption{Creation of Smaller ORQA Variants through Knowledge Distillation. The original ORQA model, teacher, supervises progressively smaller students, enabling efficient deployment in resource-constrained surgical environments.}
\label{fig:distillation}
\end{figure}

\subsection*{Training Procedure}
The ORQA model was initialized using the Qwen2-VL architecture~\cite{wang2024qwen2} for both the large language model (LLM) backbone and the image encoder, leveraging its pretrained vision-language capabilities. Modality-specific encoders, CLIP~\cite{radford2021learning}, Point Transformer V3~\cite{wu2024point}, CLAP~\cite{elizalde2023clap}, were either pretrained and frozen (e.g., CLAP) or trained end-to-end (e.g., Point Transformer). We employed Low-Rank Adaptation (LoRA)~\cite{hu2021lora} to fine-tune the LLM efficiently, reducing computational overhead while preserving pretrained knowledge. The model was trained on 1 million QA pairs using cross-entropy loss, optimized with the Adam optimizer, a learning rate of $1 \times 10^{-4}$, and a batch size of 4 for one epoch. To improve robustness, we applied modality dropping, where each modality had a 50\% chance of exclusion per sample, and modality mixing, where modalities were swapped with those from samples with similar scene graphs. These techniques ensured generalization across varied OR scenarios. To enhance conversational fluency, we generated 43,408 additional QA pairs using the non-fine-tuned Qwen2-VL model to produce scene summaries and surgical descriptions from OR data. These pairs were added to the training set, improving the model’s ability to long-form descriptions of the scene. The ORQA-Temp variant incorporates temporal modeling using memory scene graphs~\cite{ozsoy2023,ozsoy2024oracle, ozsoy2025mmor}, capturing short-term (recent triplets, e.g., (surgeon, drill, holding)) and long-term (unique triplets in order of first occurrence) context. Training was performed on a single NVIDIA A40 GPU using PyTorch 2.0 with CUDA 11.8, completing in approximately 7 days.

\noindent\textbf{Distillation into Smaller Variants.} To support deployment in resource-constrained surgical environments, we used knowledge distillation~\cite{hinton2015distilling} to compress ORQA into three smaller variants, ORQA-Dist-L, ORQA-Dist-M, and ORQA-Dist-S, optimized for varying speed and memory requirements. In knowledge distillation, a teacher model supervises a student model via KL divergence loss, defined as:

\begin{equation}
\mathcal{L} = KL\left( \sigma\left(\frac{z_t}{T}\right) || \sigma\left(\frac{z_s}{T}\right) \right) \cdot T^2
\label{eq:distillation_loss}
\end{equation}

where $\sigma$ is softmax, $z_t$ and $z_s$ are the output logits of the teacher and student models, $T$ is a hyperparameter used for softening the outputs, and KL is the KL divergence, defined as:

\begin{equation}
KL(P || Q) = \sum_i P(i) \log \left( \frac{P(i)}{Q(i)} \right)
\label{eq:kl}
\end{equation}

We trained smaller student models by iteratively reducing the size of the ORQA model. To initialize each student, we cropped the weights from the top left corner of the teacher’s weight matrices, preserving a portion of the learned representations. As visualized in~\Cref{fig:distillation}, starting with a student model matching ORQA’s size, we progressively reduced its hidden dimensions during training, with each subsequent student initialized from the weights of the previous one. This process enables efficient models suitable for real-time applications on hardware with limited computational resources.

\subsection*{Evaluation Procedure}
The ORQA model, its distilled variants, and all the other baselines were evaluated on the ORQA benchmark’s test split to assess performance across 23 tasks. The evaluation used a custom ORQA Score metric, detailed statistical analysis, and standardized protocols to ensure robust comparisons. Three proprietary generalist models, GPT-4o, GPT-4.1, Gemini 2.5 Flash were tested via their APIs in a zero-shot setting. Furthermore, we tested the base model we are starting from, Qwen2-VL (2B)~\cite{wang2024qwen2} and its bigger 7B variant. For all zero-shot experiments, Task-specific prompts provided question formats, expected outputs, and OR context (e.g., “List tools in use as a comma-separated string”). A statistical baseline, predicting the most frequent answer per task, was also included for reference.

\noindent\textbf{ORQA Score.} The ORQA Score (0–1) was computed per task with task-specific rules: 
\begin{itemize}
    \item \textit{People Counting}: 1.0 for exact matches, 0.5 if off by one, 0.0 otherwise.
    \item Set-based tasks (e.g., \textit{Role Detection}, \textit{Tool Detection}): Intersection over Union (IoU).
    \item Single-label tasks (e.g., \textit{Is Completed}, \textit{Sterility Breach Detection}): 1.0 if correct, 0.0 otherwise.
    \item Relative distance tasks (e.g., \textit{3D Distance}, \textit{Estimate Status}): 1.0 for errors $<10\%$, 0.5 for $<25\%$, 0.0 beyond.
    \item \textit{2D Detection}: IoU thresholds (1.0 for $\geq0.75$, 0.75 for $\geq0.5$, 0.5 for $\geq0.25$, 0.25 for $\geq0.125$).
    \item \textit{Scene Graph Generation}: Macro F1 score.
    \item \textit{Sorted Entity Detection}: Normalized Levenshtein distance.
    \item \textit{Monitor Text OCR}: BLEU-1 score.
\end{itemize}

Scores were averaged per task, per dataset, and overall to produce a comprehensive metric. Thus the ORQA score reflects holistic understanding rather than performance on a single task. We compute 95\% confidence intervals for the ORQA Score and all per-task scores using nonparametric bootstrapping over the test set (1,000 resamples).

\subsection*{Ethics Statement}
The ORQA benchmark is based on publicly available datasets (MVOR~\cite{mvor}, 4D-OR~\cite{ozsoy20224d}, EgoSurgery~\cite{fujii2024egosurgery}, MM-OR~\cite{ozsoy2025mmor}), containing no protected health information or human subject data, rendering it exempt from Institutional Review Board (IRB) oversight. ORQA’s capability for local inference ensures compatibility with hospital privacy policies and national data sovereignty regulations, minimizing risks of external data transfer. ORQA is trained and evaluated entirely on research datasets and is not intended for clinical deployment without further validation. While the benchmark reflects real surgical challenges, it does not model all clinical variables, nor does it replace expert medical judgment. As with all foundation models, the potential for bias, overfitting, and spurious correlations exists. We mitigate this risk by transparently releasing the full benchmark and evaluation code upon acceptance, enabling reproducibility and independent auditing. Future work will explore fairness across institutions, surgeons, and procedure types.

\subsection*{Data Availability} 
The MVOR, 4D-OR, EgoSurgery, and MM-OR datasets, are already publicly available. Our curated question answering dataset and benchmark will be made available upon acceptance.

\subsection*{Code Availability} 
The ORQA model implementation, training scripts, and evaluation code will be released on a public GitHub repository upon acceptance, ensuring reproducibility and open access.

\subsection*{Acknowledgements} 
The authors thank all members of the Chair for Computer Aided Medical Procedures and Augmented Reality (CAMP) at the Technical University of Munich for their feedback and support during the development of this work. No external funding was received for this study.

\subsection*{Author Contributions} 
E.Ö. conceived the study, collected and curated the dataset, developed the models, and led the implementation and evaluation. C.P. contributed to conceptual development, figure design, and manuscript writing. D.B.H., K.Y., and M.K. supported the writing process and participated in high-level discussions. N.N. supervised the project and provided feedback on the overall study design and manuscript preparation. All authors reviewed and approved the final manuscript.

\bibliographystyle{plain} % or unsrtnat, abbrvnat, etc.

\bibliography{sn-bibliography}

\end{document}